\documentclass[conference]{IEEEtran}
\IEEEoverridecommandlockouts
\usepackage{amsmath,amssymb,amsfonts}
\usepackage{cite}
\usepackage{algorithmic}
\usepackage{graphicx}
\usepackage{textcomp}
\usepackage{xcolor}
\usepackage{subfiles}
\usepackage{adjustbox}
\usepackage{multirow}
\usepackage{hyperref}
\def\BibTeX{{\rm B\kern-.05em{\sc i\kern-.025em b}\kern-.08em
    T\kern-.1667em\lower.7ex\hbox{E}\kern-.125emX}}

\begin{document}

\title{mmGAT: Pose Estimation by Graph Attention with Mutual Features from mmWave Radar Point Cloud\\}

\author{\IEEEauthorblockN{Masud Abdullah Al$^{\text{1}}$, Xintong Shi$^{\text{2}}$}
\IEEEauthorblockA{\textit{Graduate School of Science and Technology} \\
\textit{Keio University}\\
Kanagawa, 223-8522, Japan}
\and
\IEEEauthorblockN{Bouazizi Mondher$^{\text{3}}$, Tomoaki Ohtsuki$^{\text{4}}$}
\IEEEauthorblockA{\textit{Faculty of Science and Technology} \\
\textit{Keio University}\\
Kanagawa, 223-8522, Japan}
\and
Email: $^{\text{1}}$masud@ohtsuki.ics.keio.ac.jp, $^{\text{2}}$shixintong@ohtsuki.ics.keio.ac.jp, $^{\text{3}}$bouazizi@ohtsuki.ics.keio.ac.jp, \\$^{\text{4}}$ohtsuki@keio.jp.
}

\maketitle

\begin{abstract}

Pose estimation and human action recognition (HAR) are pivotal technologies spanning various domains. While the image-based pose estimation and HAR are widely admired for their superior performance, they lack in privacy protection and suboptimal performance in low-light and dark environments. This paper exploits the capabilities of millimeter-wave (mmWave) radar technology for human pose estimation by processing radar data with Graph Neural Network (GNN) architecture, coupled with the attention mechanism. Our goal is to capture the finer details of the radar point cloud to improve the pose estimation performance. To this end, we present a unique feature extraction technique that exploits the full potential of the GNN processing method for pose estimation. Our model mmGAT demonstrates remarkable performance on two publicly available benchmark mmWave datasets and establishes new state of the art results in most scenarios in terms of human pose estimation. Our approach achieves a noteworthy reduction of pose estimation mean per joint position error (MPJPE) by 35.6\%  and PA-MPJPE by 14.1\% from the current state of the art benchmark within this domain.

\end{abstract}

\begin{IEEEkeywords}
mmWave, graph attention, graph neural network, pose estimation
\end{IEEEkeywords}

\section{Introduction}

Radar is a century-old technology widely used for long range object detection, vehicle navigation, motion detection, trajectory estimation, etc. After recent introduction of millimeter wave (mmWave) radar, it has grown in popularity in many research fields. mmWave radars are being successfully used in pose estimation \cite{b1, b2, b4, b6, b35, b38, b41}, human action recognition \cite{b3, b5, b39}, heartbeat detection \cite{b15, b16, b17, b18}, rehabilitation \cite{b8, b9}, human vitality analysis \cite{b10}, etc. for its relatively higher resolution.

In particular, we are interested in detecting human poses and activities from the radar signals. Similar to the works in \cite{b1, b2, b4, b6, b35, b39}, our approach involves using deep learning to estimate human pose. In \cite{b1, b2, b4}, An \textit{et al.} used a Convolutional Neural Networks (CNN)-based pose estimation model and released two public datasets, the Mars dataset and the mRI dataset. In \cite{b4}, they introduced an efficient technique to fuse the point cloud of consecutive radar frames to increase data density. Though the previous works used several features captured by the radar sensors, they neglected the inclusion of essential information, such as the mutual relationship of radar points in the point cloud. Also, while sorting radar points along spatial dimension and stacking them in the image frame, closer points do not remain close and thus spatial coherence information is mostly lost. In \cite{b35}, Sengupta \textit{et al.} introduced another CNN-based pose estimation model using mmWave radar data, but they neglected the internal relationship between a pair of radar points in the point cloud. Another work was proposed in \cite{b38} by Sengupta \textit{et al.} demonstrating a Sequence-to-Sequence (seq2seq) modeling \cite{b40} approach for pose estimation and achieving high precision. However, we noticed two limitations in this approach, including the limitation of using only point cloud coordinate locations as features and the limitation of the proposed cube voxelization method.
% First, it throws away important features such as Doppler velocity, intensity, and also the internal relationships among the radar points. Second, the cube voxelization approach is less scalable while achieving high precision even in moderate resolution. 

% In \cite{b1} and \cite{b2}, two public datasets have been released, Mars dataset \cite{b1} and mRI dataset \cite{b2}.

As the radar point cloud is a set of points in 3D coordinates, we noticed that a graph can represent them. Recent progress in deep learning research has prominently featured work on graph data, among which the Graph Neural Network (GNN) stands out as a key development. GNN is very popular in many fields such as drug discovery \cite{b19, b21}, protein synthesis \cite{b22}, recommendation systems \cite{b23}, human tracking over multiple camera networks \cite{b24}, and so on. Furthermore, an interesting work was proposed in \cite{b39} by Gong \textit{et al.} which used GNN model to process radar data points for HAR.

For pose estimation, we focus not only on the features of the points themselves but also on conceptually important details of the radar point cloud such as the distance between a pair of points, the direction from one point to another, the relative Doppler velocity, and the relative reflective intensity etc., which we refer to as mutual features. To process such information, a robust and scalable technique is to utilize GNN. Specifically, we use Graph Attention (GAT) \cite{b25} to process both point features and mutual features and combine them to express the radar point cloud by an efficient feature representation. 

The main contributions of this work lie in the data processing and modeling stages.
\begin{enumerate}
    \item The point cloud of one frame is assumed as a directed graph \(G(V,E)\) where \(V\) is a set of all radar data points in a frame and \(E\) indicates the set of connections between all combinations of pair of points. To the best of our knowledge, we are the first to fully use the potential of radar point clouds by processing through GNN in a scalable way.
    \item A novel feature processing mechanism is introduced in this paper which can capture finer details among radar data points using point-to-point distances, directions along three spatial axes between two points, relative Doppler velocity, and relative reflective intensity. We indicate them as edge features or mutual features. The attributes of each radar data point, 3D location, Doppler velocity, and reflective intensity, are considered node features.
    \item We successfully utilized GAT \cite{b25}, a type of GNN model to combine both node features and edge features for generating better feature representations.
\end{enumerate}

The rest of the paper is organized as follows. We start by briefly introducing some related works. We discuss our model in detail in Section III. In Section IV, we explain our experimental setup. In Section V we discuss the experimental results. Finally, we conclude our work in Section VI.

\section{Related Work}

Recently, mmWave radar has grown in popularity for its relatively high resolution which can separate two objects as close as several millimeters apart. It is known as millimeter wave radar or mmWave radar for its wavelength being in the millimeter range. It follows the same principle as long range radars in utilizing Doppler frequency shift to detect moving and static objects. mmWave radar creates radio wave in the range between 77 GHz to 80.2 GHz \cite{b1}. It generates a chirp signal by linearly increasing its frequency within this range so that it can detect static objects. A single radar frame contains many chirp signals which form a Sawtooth wave in frequency domain. It also integrates multiple transmitter and receiver antennas leading to a higher number of virtual arrays which gives mmWave radars more detection precision. 

Pose estimation is one of the most interesting challenges for radar as the radar point clouds are directly related to human movements and their motion relative to the radar. In \cite{b1, b2, b4}, An \textit{et al.} used a CNN-based modeling approach to process the radar point cloud data. They restructured radar point cloud data into a 3D array with 5 channels. They fitted each point within an image pixel and created an image-frame of size (\(14\times14\times5\)) by arranging and adding zero padding 192 radar points from a single radar frame. They also used several other techniques proved to be useful in pre-processing stages such as sorting data in spatial axes, fusing points from consecutive frames to increase data density.

Another work proposed in \cite{b6} by Yu \textit{et al.} used a two-stage pose estimation model. They used a human localization model to predict the heatmap of the location of subject and then cropped the localization heatmap to use closer details for pose estimation of each person. One of the most influential works on radar-based pose estimation was described in \cite{b35}, where Sengupta \textit{et al.} introduced a CNN-based pose estimation model from mmWave radar data which created two 3D feature vectors from each radar data point, one using (depth, azimuth, intensity) values and the other using (depth, elevation, intensity) values. Afterwards, they considered each vector as an image pixel, stacked all pixels from all radar points of a frame into an array of size (\(16\times16\)) and filled rest of the pixels by zero padding. Thus, they created two (\(16\times16\times3\)) image arrays and processed them using CNN. Another work was proposed in \cite{b38} by Sengupta \textit{et al.} demonstrating seq2seq modeling \cite{b40} approach for pose estimation which converted voxelized radar points into tokens and followed the seq2seq approach used in Natural Language Processing (NLP) to first process the data through an encoder block and then predicted voxelized pose keypoints at the output in a recurrent manner. Finally, the output was devoxelized to obtain pose keypoint locations.

In \cite{b39}, Gong \textit{et al.} used a GNN model to process radar data points for human activity recognition. The GNN used all available radar point cloud features such as point coordinates, Doppler velocity, intensity, angle, etc. as node features and computed edge features from node features using Multi-Layer Perceptron (MLP) layers. They combined the node and edge features using MLP and some other cascaded operations.

One of the initial and exemplary works on radar for action detection is described in \cite{b5}, where 3 radar antennas were used to capture different motions and tried to estimate the action. One of the initial works using deep learning and machine learning was proposed in \cite{b3}. They introduced an mmWave dataset consisting of 5 activities. They converted the radar point cloud into voxels and created sequential frame data to predict activities following a sequential modeling approach. They used Bi-Directional Long-Short Term Memory (LSTM), MLP, Support Vector Machine (SVM) etc. technologies to predict human activity and achieved impressive results. Another work in \cite{b27} used a similar voxelization approach for detecting sitting direction arrangements using sequential modeling.

In this paper, in terms of data pre-processing, there are notable similarities adapted from \cite{b1} and \cite{b4}, but we design a robust point cloud mutual feature extraction method and demonstrate how to utilize these features using GAT. Another work \cite{b28} utilized GNN for pose estimation from mmWave radar data. However, the authors only used it for refining the already predicted pose. Another work proposed in \cite{b39} by Gong \textit{et al.} used GNN for radar point cloud. However, there are significant differences between our modeling approach and theirs. They did not compute any mutual edge features rather preferred to calculate edge features from another neural network block. On the other hand, we try to use more conceptually insightful features. Furthermore, we use a graph attention module to combine node and edge features.

Our objective is to capture finer features in the point cloud and further improve the accuracy of human pose estimation. Hence, we collected features inside point cloud such as distance between a pair of points, direction from one point to another, relative velocity between two points, etc. To process such data, conventional neural network architectures such as Fully Connected Network (FCN), CNN, LSTM, etc. are not good match. Nevertheless, GNN is designed to handle such data where there are inter-relationships among data points. Since radar data points are reflected from the human body structure, the points must be related to each other based on skeleton structure. That is what motivated us to use GNN. Particularly, GAT has the ability to process both the node feature and the mutual edge feature and produce a combined representation of the entire graph structure. This is why we decided to use GAT.

\section{Proposed Method}

In our model, we designed the whole model as a graph-based solution. We will discuss them as follows.

\subsection{Model Definition}

Suppose, a radar data frame \(i\) contains \(n_i\) data points. 
We construct a directed graph \(G_i(V, E)\) for frame \(i\), where \(V=\{p_1, p_2,\cdots, p_{n_i}\}\) is a set of nodes and \(E=\{e_{1,2}, e_{2,1},\cdots, e_{n_i,n_i-1}\}\) is a set of directed edges.
We connect the nodes in a directed way so that there can be a connection from \(p_1\) to \(p_2\) denoted by \(e_{1,2}\), but there may not be a connection from \(p_2\) to \(p_1\). 
Thus, \(E\) will not contain all permutation of edges. We take a target point \(p_j\) and connect it with the \(K\) nearest points based on Euclidean distance. This makes the model scalable for high density point clouds. Let the neighboring set of \(p_j\) to be \(N(p_j)\). 
Here is an example, \(N(p_j)\) = \(\{p_3, p_7, p_8,\cdots, p_{n_i-5}\}\). Hence, \(e_{p_j,p_3}, e_{p_j,p_7},\cdots, e_{p_j,p_{n_i-5}}\) will be included in \(E\). 
For terminology purposes, ``node" and ``point" are interchangeably used to indicate a single radar data point.

\subsection{Feature preparation}

In both MARS \cite{b1} and mRI \cite{b2} datasets, for each radar data point, we get 5 values which are \(x, y, z\) location values, Doppler velocity (\(v\)), and intensity (\(I\)). Thus, every node from \(V\) has these features. We consider them as node feature \(f_{p_j}\). We calculate some mutual features among all pairs of data points. We consider them as edge features. Edge features between \(p_1\) and \(p_2\) are indicated as \(g_{e_{1,2}}\).

There are 6 mutual/edge features which are
\begin{enumerate}
    \item The Euclidean distance  \(||p_1 - p_2||\),
    \item The angles from \(p_1\) to \(p_2\) respective to the \(x, y, z\) axes,\\
    \((p_2-p_1) / ||p_2-p_1||\). If the denominator is \(0\), the angle is considered as \(0\),
    \item The relative velocity  \(v_{p_2}-v_{p_1}\),
    \item The relative intensity  \(I_{p_2}-I_{p_1}\).
\end{enumerate}
Here \(||.||\) indicates \(L_2\) norm. So, \(f_{p_1}\) is a 5 dimensional vector and \(g_{e_{1,2}}\) is a 6 dimensional vector. While calculating features, we calculated edge feature values for all possible permutations of node pairs.
\begin{figure}[htbp]
\centering\includegraphics[width=0.45\textwidth]{../images/features.png}
\caption{Feature description.}
\label{fig-feature}
\end{figure}
However, graph data are stored and used in the GNN model based on their neighboring sets. 

Though it does not provide additional value to the graph processing algorithm, we still used radar point sorting in spatial axes similar to \cite{b1}. We also used fusion of radar points from consecutive frames to increase data density adapted from \cite{b4} by stacking data points of three consecutive frames. Our model was tested with and without fusing data points of consecutive frames. We found that increasing data density provides positive response to our feature processing approach and gives good performance boost.

\subsection{Model Architecture}

\subsubsection{Edge Feature Processing Block (\(h_{edge}\))}

We hypothesize that edge features contain useful information about the graph data. Moreover, both the node features and the edge features need to be processed, each, in an appropriate way. Therefore, we process the edge feature vectors using the edge feature processing block. 

Our edge feature processing block \(h_{edge}\) consists of 3 FCN layers (popularly known as Dense layer or Linear layer). The 2nd and 3rd layers are followed by Rectified Linear Unit (ReLU) \cite{b37} activation function. Each layer has 64 units. Thus, we get the processed edge feature \(X_{p_{j_{edge}}}\) between a target point \(p_j\) and all its neighbors \(N(p_j)\) as follows:

\begin{equation}
    X_{p_{j_{edge}}} = h_{\text{edge}}(g_{e_{p_j,p_k}}) , p_k \in N(p_j).
    \label{edge-feature}
\end{equation}

\subsubsection{Graph Attention Block (\(h_{gat}\))}

Next, we process the node features and combine them with the processed edge features \(X_{p_{j_{edge}}}\) in GAT computation. GAT computation calculates an attention weight for each neighboring node to decide how much importance should be given to which neighbor. It processes the target node feature with the neighboring nodes via a series of linear transformations. It follows the equations below and finally calculates an updated feature representation of the target node \(p_j\) as \(X_{p_{j_{node}}}\) 

\begin{equation}
    X_{p_{j_{node}}} = h_{gat}(f_{p_j}, N(p_j), X_{p_{j_{edge}}}).
    \label{node-feature}
\end{equation}
We can break down Eq. \ref{node-feature} into two parts. \(h_{gat}\) symbolizes accumulation of these two parts. Firstly, we calculate the attention weight \(\alpha_{j,k}\) as follows:
\begin{equation}
    \small
    \alpha_{j,k} = \frac{\exp(a(W^T([\Theta(f_{p_j}), \Theta(f_{p_k}), \Theta_e(X_{p_{j_{edge}}}[k])])))}
    { \sum_{p_k \in N(p_j)} \exp(a(W^T([\Theta(f_{p_j}), \Theta(f_{p_k}), \Theta_e(X_{p_{j_{edge}}}[k])])))}.
    \label{attention-weight}
\end{equation}
\begin{figure}[htbp]
\centering\includegraphics[width=.5\textwidth]{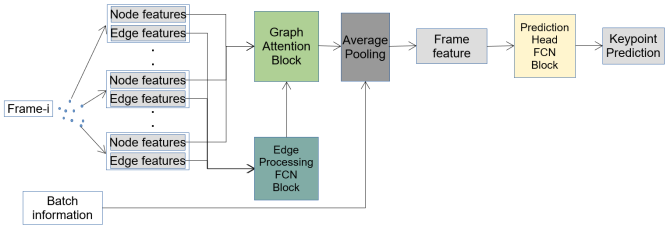}
\caption{The architecture of the mmGAT model.}
\label{fig-model}
\end{figure}
Afterwards, we use \(\alpha_{j,k}\) to calculate the updated node feature
\begin{equation}
    X_{p_{j_{node}}} = \alpha_{j,j}*\Theta(f_{p_j}) + \sum_{p_k\in N(p_j)}\alpha_{j,k}*\Theta(f_{p_k}).
    \label{node-feature-detail}
\end{equation}
Here \(W\), \(\Theta\) and \(\Theta_e\) are learnable parameters, \(a\) is the Leaky ReLU activation function and \([\Theta(f_{p_j}), \Theta(f_{p_k}), \Theta_e(X_{p_{j_edge}}[k])]\) is a concatenation operator. After processing the node and edge features, we get the processed node features for the target node \(p_j\). The process is repeated for all the nodes in \(V\).

We use a dropout layer with the ratio of 0.5 inside the GAT layer. The number of neurons for a GAT layer is 128. We use 4 consecutive GAT layers to process node features each one followed by a ReLU activation function except the last layer.

\subsubsection{Feature Aggregation by Pooling}

Now that we have the node level representation vectors, we have to combine them to create a graph level representation vector. In this regard, we can use an aggregation function such as average, minimum, maximum etc. pooling operations. In our case, we tried a few options and found that average pooling works the best. So, we take all the node features and average them to get a graph level representation feature vector as follows,

\begin{equation}
    m_i = \frac{1}{n_i} \sum_{j=1}^{n_i} X_{p_{j_{node}}}.
    \label{aggregation}
\end{equation}

While processing data in a batch of multiple frames, we also need information about which point belongs to which radar frame to calculate this aggregation.

\subsubsection{Pose Prediction Head (\(h_{head}\))}

Finally we get a vector representation of the radar point-cloud as \(m_i\). Now we process this feature vector using pose prediction block \(h_{head}\) to get final pose estimation. \(h_{head}\) consists of 5 fully connected layers each followed by a ReLU activation function except the last layer. Each layer contains 256 units except the last layer which contains same number of units as 3 times the number of total keypoints (for the mRI dataset \(51=3\times17\), 17 is the number of total keypoints). Thus, as shown in Eq.~\ref{pose-prediction} we get the pose estimation output as a form of regression computation in a flattened vector.

\begin{equation}
    X_{pose} = h_{head}(m_i).
    \label{pose-prediction}
\end{equation}

\section{Experimental Setup}

For our experimentation, we selected two public mmWave pose estimation datasets which are MARS \cite{b1} and mRI \cite{b2} datasets. Since our main focus is to detect human pose from radar data, we only utilize radar data from mRI dataset. Also, for fair comparison of the results in
these datasets, we implement a similar frame-by-frame pose estimation model as described in \cite{b1} and \cite{b2}. For Mars dataset, the data comes with train, validation and test sets with a split ratio of 60\%, 20\%, 20\% respectively. For evaluating the model performance, they used popular metrics MAE (Mean Absolute Error) and RMSE (Root Mean Squared Error). We also used the same evaluation metrics. For the loss calculation for training the model, we used a simple MSE loss, which seems appropriate for the task and also for the evaluation metrics. For the mRI dataset, according to \cite{b2}, they split the dataset into total of 4 modes. They took two data splitting approaches which are the random splitting (S1) and the subject-wise splitting (S2). Again, for each of these splitting modes, they selected two types of activities which are protocol-1 (P1) consisting of 12 activities and protocol-2 (P2) consisting of 10 activities. Thus, they got a total of four combinations such as S1P1, S1P2, S2P1 and S2P2. We followed exactly the same approach for keeping the result comparison fair and consistent. For dataset splitting, we used 80\%, 20\% splitting between training and test set respectively same as the mRI paper \cite{b2}. In our case, among 20 subjects participated in the experiment, subjects for test dataset were chosen as (1, 4, 7, 15) by random selection for the S2 mode. We created S2P1 and S2P2 from this arrangement.

In the mRI paper \cite{b2}, they used MPJPE and Procrustes Analysis MPJPE (PA-MPJPE) \cite{b30} as evaluation metrics. For proper result analysis, we also used the same evaluation metrics. For loss calculation for training the model, we used the MPJPE loss which is appropriate for the task and also for the evaluation metrics.

We have already described the complete model architecture in Section III. The number of neighboring nodes \(K\) is 20 in our experiment as it seemed to be enough memory efficient and at the same time provides reliable performance for both the Mars and the mRI datasets. For the mRI dataset, we have used a batch size of 128 (128 radar frame graphs), we trained the models for 250 epochs for all the scenarios, we used the Adam optimizer for training the model with back-propagation, used the lambda training scheduler with a reduction factor of 0.995 (after every epoch, the learning rate is multiplied by 0.995), the primary learning rate was 0.001. For the MARS dataset, we have used a batch size of 32, we trained the models for 160 epochs, the primary learning rate was 0.001 with the lambda learning rate scheduler with a reduction factor of 0.995.

\section{Results and Discussion}

On the Mars \cite{b1} dataset, we are reporting the model performance in terms of MAE and RMSE calculated from the predicted pose keypoints from our model and the ground truth 3D pose keypoints. The average MAE and RMSE scores along \(x, y, z\) axes and the overall average MAE and RMSE scores are reported in Table I.

\begin{table}[htbp]
\caption{Results on Mars \cite{b1} dataset.}
\begin{center}
\begin{adjustbox}{width=1\columnwidth, left}
\begin{tabular}{|c|cc|cc|cc|cc|}
\hline
\multirow{2}{*}{\textbf{Model}} & \multicolumn{2}{|c|}{\textbf{X-axis (cm)}} & \multicolumn{2}{|c|}{\textbf{Y-axis (cm)}} & \multicolumn{2}{|c|}{\textbf{Z-axis (cm)}} & \multicolumn{2}{|c|}{\textbf{Average (cm)}} \\
\cline{2-9}
 & MAE & RMSE & MAE & RMSE & MAE & RMSE & MAE & RMSE \\
\hline
\hline
Mars \cite{b1} & 6.99 & 9.79 & 4.07 & 5.56 & 6.54 & 8.94 & 5.87 & 8.10 \\
Fast scalable \cite{b4} & 4.2 &  & 2.5 &  & 4.4 &  & 3.6 &  \\
mmGAT model (ours) & 3.593 & 6.277 & 2.788 & 4.121 & 3.878 & 6.561 & 3.42 & 5.653 \\
\hline
\end{tabular}
\label{tab1}
\end{adjustbox}
\end{center}
\end{table}

\begin{table}[htbp]
\begin{center}
\caption{Results on mRI \cite{b2} dataset.}
\begin{adjustbox}{width=1\columnwidth, left}
\begin{tabular}{|c|c|c|cc|}
\hline
Dataset formation & Protocol & Model & MPJPE (mm) & PA-MPJPE (mm) \\
\hline
\hline
\multirow{4}{*}{S1 (random split)} & \multirow{2}{*}{P1} & mRI paper & 163.3 & 94.1 \\
 &  & mmGAT model & 82.786 & 70.934 \\
 \cline{2-5}
 & \multirow{2}{*}{P2} & mRI paper & 125.1 & 74.1 \\
 &  & mmGAT model & 66.045 & 48.547 \\
\hline
\multirow{4}{*}{S2 (subject-wise split)} & \multirow{2}{*}{P1} & mRI paper & 186.6 & 97.3 \\
 &  & mmGAT model & 125.514 & 98.972 \\
 \cline{2-5}
 & \multirow{2}{*}{P2} & mRI paper & 126.6 & 75.0 \\
 &  & mmGAT model & 109.865 & 75.479 \\
\hline
\end{tabular}
\end{adjustbox}
\end{center}
\end{table}
Mars \cite{b1} paper did not utilize the data fusion technique introduced in \cite{b4} which we used in this paper. We noticed that adding data fusion of consecutive frames improved the pose estimation performance in the CNN-based model in \cite{b4} by reducing the average MAE score to 3.6 cm. When we adopted this data fusion technique, we discovered that this increased data density helps our mmGAT model to perform even better and reduces the average MAE to 3.42 cm. This is a direct indication that the usage of GAT and mutual features have accelerated the data fusion technique to achieve better point cloud feature processing.

For the mRI dataset, we evaluated our model's performance on pose estimation. We have used MPJPE and PA-MPJPE \cite{b30} scores in accordance with the mRI \cite{b2} paper’s reported results. The MPJPE score was calculated after adjusting the predicted skeleton position by shifting the pelvis (mid of both hips) point to the pelvis of ground truth same as described in the mRI paper. The results are shown in Table II. As we can see, our model performed significantly better in most scenarios. We noticed that our results have relatively much smaller difference between MPJPE and PA-MPJPE scores. That means there are less rotated and poorly scaled skeletons than \cite{b2}. It led us to believe that mmGAT can explore and process the mmWave point cloud space better than the CNN-based model used in \cite{b2}. As a result, on average, we were able to reduce MPJPE score by 35.6\% and PA-MPJPE score by 14.1\%.

To understand the effect of the mutual features, we additionally trained our model on the mRI \cite{b2} dataset for each of the above four scenarios without using the mutual features. We have seen that in all the scenarios, our model can obtain lower MPJPE and PA-MPJPE scores while using mutual features. It is a strong indication that mutual features are very helpful elements of our modeling approach. A result comparison between our baseline mmGAT model (\(K=20\)) and another version of the model where mutual features are neglected (\(K=0\)) is shown in Table III. We can see that the mutual features are showing lower errors to the performance in all the scenarios. 
\begin{table}[htbp]
\begin{center}
\caption{Result comparison on mRI \cite{b2} dataset between \(K=20\) and \(K=0\).}
\begin{adjustbox}{width=1\columnwidth, left}
\begin{tabular}{|c|c|cc|cc|}

\hline
\multirow{2}{*}{Split mode} & \multirow{2}{*}{Protocol} & \multicolumn{2}{|c|}{without mutual features (K=0)} & \multicolumn{2}{|c|}{with mutual features (K=20)}  \\
\cline{3-6}
 &  &  MPJPE (mm) & PA-MPJPE (mm) & MPJPE (mm) & PA-MPJPE (mm) \\
\hline
\hline
\multirow{2}{*}{S1} & P1 & 83.063 & 70.594 & 82.786 & 70.934 \\
 & P2 & 66.824 & 49.15 & 66.045 & 48.547 \\
\hline
\multirow{2}{*}{S2} & P1 & 127.738 & 100.547 & 125.514 & 98.972 \\
 & P2 & 111.977 & 77.351 & 109.865 & 75.479 \\
\hline
\end{tabular}
\end{adjustbox}
\end{center}
\end{table}
\begin{table}[htbp]
\begin{center}
\caption{Cumulative density function for the position of ground truth pose keypoints.}
\begin{adjustbox}{width=1\columnwidth, left}
\begin{tabular}{|c|c|c|c|c|c|c|c|c|c|c|c|}
\hline
 & -5 & -4 & -3 & -2 & -1 & 0 & 1 & 2 & 3 & 4 & 5 \\
\hline
x-mean & .002 & .002 & .004 & .005 & .078 & 55.058 & 97.890 & 99.989 & 99.994 & 99.995 & 99.997 \\
y-mean & 0.0 & 0.0 & .001 & .001 & 0.002 & 31.36 & 99.976 & 99.997 & 99.998 & 99.999 & 99.999 \\
z-mean & .005 & .006 & .007 & .008 & .009 & .013 & 1.186 & 3.103 & 99.365 & 99.974 & 99.983 \\
\hline
\end{tabular}
\end{adjustbox}
\end{center}
\end{table}
\begin{table}[ht!]

\caption{Results on mRI \cite{b2} dataset after denoising.}
\begin{tabular}{|c|c|cc|}

\hline
Dataset formation & Protocol & MPJPE (mm) & PA-MPJPE (mm) \\
\hline
\hline
\multirow{2}{*}{S1 (random split)} & P1 & 76.406 & 55.159 \\
 & P2 & 66.669 & 48.874 \\
\hline
\multirow{2}{*}{S2 (subject-wise split)} & P1 & 115.793 & 77.855 \\
 & P2 & 110.258 & 74.46 \\
\hline
\end{tabular}
\end{table} 

We found that there are some frames in the mRI dataset where the ground truth pose keypoints are noisy and much far away from the skeleton. Table IV shows the cumulative density function (CDF) for the position of ground truth pose keypoints in between \(0 \sim 100\%\). Here the values from -5 m (meter) to +5 m are coordinate position values. x-mean at -5 m is 0.002 meaning that along x-axis, 0.002\% samples have pose keypoint positions below -5 m on average for all subjects in mRI dataset. We can consider the table as average cumulative density function of ground truth pose keypoint locations for all the subjects along all three axes. As we can see, outside (-5 m, +5 m) range, only a few samples are present. However, we noticed that some samples have very small and very large values which affect the overall results to some extent. Hence, we tried to apply volume restriction based denoising by setting a validity range for pose keypoints locations by (-5 m, +5 m) along all three axes in the experiment below. We removed all the radar frames from train, validation, and test datasets if any ground truth pose keypoint lies outside this volume. We checked the statistics of the removed points and found that this volume restriction removed less than 0.01\% data from all the datasets. Yet, it has a significant effect in the model performance. The results after denoising is shown in Table V.

It is worth noting that the data fusion technique for consecutive radar frames from \cite{b4} is quite effective in our data processing method. We tried to use the Mars \cite{b1} dataset without using fusion of data points. We noticed that under this condition, the mmGAT model performs almost similar to the CNN based model sometimes marginally better, sometimes marginally worse. After introducing data fusion processing, the mmGAT model performs way better, even better than the reported performance in \cite{b4} which indicates that the performance boost is a combined effect of both the data fusion and mmGAT feature processing techniques.

Finally, we would like to address some limitations of our work. In this work, the proposed model is specifically designed for predicting the pose keypoints of a single individual. Consequently, it lacks the capability to predict pose keypoints for multiple individuals when they are present in the same scene. Furthermore, applying our model becomes challenging when the individual is not fully within the radar monitoring range. For real-world applications, our model currently faces a limitation. If a frame contains no people but does have points reflected by objects, the model will inaccurately predict human body pose keypoints, which is inappropriate. Recognizing these constraints, our future research aims to refine the method and address these issues.

\section{Conclusion and Future Works}

In this paper, we proposed a novel mutual feature extraction method for mmWave point cloud in a graph structure. We coupled it with GAT which processed the features in an efficient way. mmGAT shows great promise in GNN-based feature processing for mmWave radar. We empirically showed that the mutual features are helpful in achieving lower MPJPE and PA-MPJPE scores. Both node and edge features provide a rich feature representation for pose estimation. mmGAT raises the state of the art to a new standard in all the evaluation metrics. We envision its usefulness in more engaging applications such as human action recognition, vitality analysis, heartbeat detection, etc.

This work was supported by JST ASPIRE Grant Number JPMJAP2326, Japan.

\bibliographystyle{IEEEtran}
\bibliography{ref.bib}

@String(ICASSP=	{ICASSP})

@article{b1,
author = {An, Sizhe and Ogras, Umit Y.},
title = {MARS: MmWave-Based Assistive Rehabilitation System for Smart Healthcare},
year = {2021},
issue_date = {October 2021},
volume = {20},
number = {5s},
issn = {1539-9087},
url = {https://doi.org/10.1145/3477003},
doi = {10.1145/3477003},
journal = {ACM Trans. Embed. Comput. Syst.},
month = sep,
articleno = {72},
numpages = {22}
}

@inproceedings{b2,
 author = {An, Sizhe and Li, Yin and Ogras, Umit},
 booktitle = {Advances in Neural Information Processing Systems},
 editor = {S. Koyejo and S. Mohamed and A. Agarwal and D. Belgrave and K. Cho and A. Oh},
 pages = {27414--27426},
 publisher = {Curran Associates, Inc.},
 title = {mRI: Multi-modal 3D Human Pose Estimation Dataset using mmWave, RGB-D, and Inertial Sensors},
 url = {https://proceedings.neurips.cc/paper_files/paper/2022/file/af9c9c6d2da701da5a0acf91ec217815-Paper-Datasets_and_Benchmarks.pdf},
 volume = {35},
 year = {2022}
}

@inproceedings{b3,
author = {Singh, Akash Deep and Sandha, Sandeep Singh and Garcia, Luis and Srivastava, Mani},
title = {RadHAR: Human Activity Recognition from Point Clouds Generated through a Millimeter-Wave Radar},
year = {2019},
isbn = {9781450369329},
publisher = {Association for Computing Machinery},
address = {New York, NY, USA},
url = {https://doi.org/10.1145/3349624.3356768},
doi = {10.1145/3349624.3356768},
booktitle = {Proceedings of the 3rd ACM Workshop on Millimeter-Wave Networks and Sensing Systems},
pages = {51–56},
numpages = {6},
location = {Los Cabos, Mexico},
series = {mmNets'19}
}

@inproceedings{b4,
author = {An, Sizhe and Ogras, Umit Y.},
title = {Fast and Scalable Human Pose Estimation Using MmWave Point Cloud},
year = {2022},
isbn = {9781450391429},
publisher = {Association for Computing Machinery},
address = {New York, NY, USA},
url = {https://doi.org/10.1145/3489517.3530522},
doi = {10.1145/3489517.3530522},
booktitle = {Proceedings of the 59th ACM/IEEE Design Automation Conference},
pages = {889–894},
numpages = {6},
location = {San Francisco, California},
series = {DAC '22}
}

@INPROCEEDINGS{b5,
  author={Okamoto, Yoshihisa and Ohtsuki, Tomoaki},
  booktitle={Proc. IEEE Int. Conf. Commun. (ICC)}, 
  title={Human activity classification and localization using bistatic three frequency {C}{W} radar}, 
  year={2013},
  volume={},
  number={},
  pages={4808-4812},
  doi={10.1109/ICC.2013.6655335}}

@INPROCEEDINGS{b6,
  author={Yu, Cong and Zhang, Dongheng and Wu, Zhi and Xie, Chunyang and Lu, Zhi and Hu, Yang and Chen, Yan},
  booktitle={ICASSP 2023 - 2023 IEEE International Conference on Acoustics, Speech and Signal Processing (ICASSP)}, 
  title={Fast 3D Human Pose Estimation Using RF Signals}, 
  year={2023},
  volume={},
  number={},
  pages={1-5},
  doi={10.1109/ICASSP49357.2023.10094778}}

@INPROCEEDINGS{b8,
  author={Takabatake, Wataru and Yamamoto, Kohei and Toyoda, Kentaroh and Ohtsuki, Tomoaki and Shibata, Yohei and Nagate, Atsushi},
  booktitle={2019 IEEE Global Communications Conference (GLOBECOM)}, 
  title={{F}{M}{C}{W} Radar-Based Anomaly Detection in Toilet by Supervised Machine Learning Classifier}, 
  year={2019},
  volume={},
  number={},
  pages={1-6},
  doi={10.1109/GLOBECOM38437.2019.9014123}}

@INPROCEEDINGS{b9,
  author={Hong, Jihoon and Tomii, Shoichiro and Ohtsuki, Tomoaki},
  booktitle={2013 IEEE 24th Annual International Symposium on Personal, Indoor, and Mobile Radio Communications (PIMRC)}, 
  title={Cooperative fall detection using Doppler radar and array sensor}, 
  year={2013},
  volume={},
  number={},
  pages={3492-3496},
  doi={10.1109/PIMRC.2013.6666753}}

@INPROCEEDINGS{b10,
  author={Yamamoto, Kohei and Toyoda, Kentaroh and Ohtsuki, Tomoaki},
  booktitle={2019 IEEE Global Communications Conference (GLOBECOM)}, 
  title={CNN-Based Respiration Rate Estimation in Indoor Environments via MIMO FMCW Radar}, 
  year={2019},
  volume={},
  number={},
  pages={1-6},
  doi={10.1109/GLOBECOM38437.2019.9013951}}

@INPROCEEDINGS{b15,
  author={Bouazizi, Mondher and Yamamoto, Kohei and Ohtsuki, Tomoaki},
  booktitle={ICC 2022 - IEEE International Conference on Communications}, 
  title={Heartbeat Detection Using 3D Lidar and MIMO Doppler Radar}, 
  year={2022},
  volume={},
  number={},
  pages={3040-3045},
  doi={10.1109/ICC45855.2022.9838339}}

@INPROCEEDINGS{b16,
  author={Yamamoto, Kohei and Endo, Koji and Ohtsuki, Tomoaki},
  booktitle={2021 IEEE Global Communications Conference (GLOBECOM)}, 
  title={Remote Sensing of Heartbeat based on Space Diversity Using MIMO FMCW Radar}, 
  year={2021},
  volume={},
  number={},
  pages={1-6},
  doi={10.1109/GLOBECOM46510.2021.9685033}}

@ARTICLE{b17,
  author={Ye, Chen and Toyoda, Kentaroh and Ohtsuki, Tomoaki},
  journal={IEEE Transactions on Biomedical Engineering}, 
  title={A Stochastic Gradient Approach for Robust Heartbeat Detection With Doppler Radar Using Time-Window-Variation Technique}, 
  year={2019},
  volume={66},
  number={6},
  pages={1730-1741},
  doi={10.1109/TBME.2018.2878881}}

@article{b18,
  title={A GAN-Based Data Augmentation Approach to Improve Heart Rate Range Classification via Doppler Radar},
  author={Yu, Danyuan and Bouazizi, Mondher and Ohtsuki, Tomoaki},
  journal={IEICE Technical Report; IEICE Tech. Rep.},
  volume={123},
  number={110},
  pages={46--51},
  year={2023},
  publisher={IEICE}
}

@article{b19,
author = {Xiong, Zhaoping and Wang, Dingyan and Liu, Xiaohong and Zhong, Feisheng and Wan, Xiaozhe and Li, Xutong and Li, Zhaojun and Luo, Xiaomin and Chen, Kaixian and Jiang, Hualiang and Zheng, Mingyue},
title = {Pushing the Boundaries of Molecular Representation for Drug Discovery with the Graph Attention Mechanism},
journal = {Journal of Medicinal Chemistry},
volume = {63},
number = {16},
pages = {8749-8760},
year = {2020},
doi = {10.1021/acs.jmedchem.9b00959},
    note ={PMID: 31408336},

URL = { 
        https://doi.org/10.1021/acs.jmedchem.9b00959
    
},
eprint = { 
        https://doi.org/10.1021/acs.jmedchem.9b00959
    
}

}

@article{b21,
title = {Graph neural networks for automated de novo drug design},
journal = {Drug Discovery Today},
volume = {26},
number = {6},
pages = {1382-1393},
year = {2021},
issn = {1359-6446},
doi = {https://doi.org/10.1016/j.drudis.2021.02.011},
url = {https://www.sciencedirect.com/science/article/pii/S1359644621000787},
author = {Jiacheng Xiong and Zhaoping Xiong and Kaixian Chen and Hualiang Jiang and Mingyue Zheng},
}

@article{b22,
  title={Fast and flexible protein design using deep graph neural networks},
  author={Strokach, Alexey and Becerra, David and Corbi-Verge, Carles and Perez-Riba, Albert and Kim, Philip M},
  journal={Cell systems},
  volume={11},
  number={4},
  pages={402--411},
  year={2020},
  publisher={Elsevier}
}

@article{b23,
title = {A deeper graph neural network for recommender systems},
journal = {Knowledge-Based Systems},
volume = {185},
pages = {105020},
year = {2019},
issn = {0950-7051},
doi = {https://doi.org/10.1016/j.knosys.2019.105020},
url = {https://www.sciencedirect.com/science/article/pii/S0950705119304277},
author = {Ruiping Yin and Kan Li and Guangquan Zhang and Jie Lu},
}

@ARTICLE{b24,
  author={Luna, Elena and SanMiguel, Juan C. and Martínez, José M. and Carballeira, Pablo},
  journal={IEEE Transactions on Circuits and Systems for Video Technology}, 
  title={Graph Neural Networks for Cross-Camera Data Association}, 
  year={2023},
  volume={33},
  number={2},
  pages={589-601},
  doi={10.1109/TCSVT.2022.3207223}}

@inproceedings{b25,
title={Graph Attention Networks},
author={Petar Veličković and Guillem Cucurull and Arantxa Casanova and Adriana Romero and Pietro Liò and Yoshua Bengio},
booktitle={International Conference on Learning Representations},
year={2018},
url={https://openreview.net/forum?id=rJXMpikCZ},
}

@Article{b27,
AUTHOR = {Wu, Jiacheng and Cui, Han and Dahnoun, Naim},
TITLE = {A Voxelization Algorithm for Reconstructing mmWave Radar Point Cloud and an Application on Posture Classification for Low Energy Consumption Platform},
JOURNAL = {Sustainability},
VOLUME = {15},
YEAR = {2023},
NUMBER = {4},
ARTICLE-NUMBER = {3342},
URL = {https://www.mdpi.com/2071-1050/15/4/3342},
ISSN = {2071-1050},
DOI = {10.3390/su15043342}
}

@InProceedings{b28,
    author    = {Lee, Shih-Po and Kini, Niraj Prakash and Peng, Wen-Hsiao and Ma, Ching-Wen and Hwang, Jenq-Neng},
    title     = {HuPR: A Benchmark for Human Pose Estimation Using Millimeter Wave Radar},
    booktitle = {Proceedings of the IEEE/CVF Winter Conference on Applications of Computer Vision (WACV)},
    month     = {January},
    year      = {2023},
    pages     = {5715-5724}
}

@article{b30,
  title={Procrustes analysis},
  author={Ross, Amy},
  journal={Course report, Department of Computer Science and Engineering, University of South Carolina},
  volume={26},
  pages={1--8},
  year={2004},
  publisher={Citeseer}
}

@ARTICLE{b35,
  author={Sengupta, Arindam and Jin, Feng and Zhang, Renyuan and Cao, Siyang},
  journal={IEEE Sensors Journal}, 
  title={mm-Pose: Real-Time Human Skeletal Posture Estimation Using mmWave Radars and CNNs}, 
  year={2020},
  volume={20},
  number={17},
  pages={10032-10044},
  doi={10.1109/JSEN.2020.2991741}}

@inproceedings{b37,
  title={Rectified linear units improve restricted boltzmann machines},
  author={Nair, Vinod and Hinton, Geoffrey E},
  booktitle={Proceedings of the 27th international conference on machine learning (ICML-10)},
  pages={807--814},
  year={2010}
}

@ARTICLE{b38,
  author={Sengupta, Arindam and Cao, Siyang},
  journal={IEEE Transactions on Neural Networks and Learning Systems}, 
  title={mmPose-NLP: A Natural Language Processing Approach to Precise Skeletal Pose Estimation Using mmWave Radars}, 
  year={2022},
  volume={},
  number={},
  pages={1-12},
  doi={10.1109/TNNLS.2022.3151101}}

@INPROCEEDINGS{b39,
  author={Gong, Peixian and Wang, Chunyu and Zhang, Lihua},
  booktitle={Proc. Int. Joint Conf. Neural Net. (IJCNN)}, 
  title={MMPoint-GNN: Graph Neural Network with Dynamic Edges for Human Activity Recognition through a Millimeter-Wave Radar}, 
  year={2021},
  volume={},
  number={},
  pages={1-7},
  doi={10.1109/IJCNN52387.2021.9533989}}

@inproceedings{b40,
 author = {Sutskever, Ilya and Vinyals, Oriol and Le, Quoc V},
 booktitle = {Advances in Neural Information Processing Systems},
 editor = {Z. Ghahramani and M. Welling and C. Cortes and N. Lawrence and K.Q. Weinberger},
 pages = {},
 publisher = {Curran Associates, Inc.},
 title = {Sequence to Sequence Learning with Neural Networks},
 url = {https://proceedings.neurips.cc/paper_files/paper/2014/file/a14ac55a4f27472c5d894ec1c3c743d2-Paper.pdf},
 volume = {27},
 year = {2014}
}

@INPROCEEDINGS{b41,
  author={Shi, Xintong and Ohtsuki, Tomoaki},
  booktitle={2023 IEEE SENSORS}, 
  title={A Robust Multi-Frame mmWave Radar Point Cloud-Based Human Skeleton Estimation Approach with Point Cloud Reliability Assessment}, 
  year={2023},
  volume={},
  number={},
  pages={1-4},
  doi={10.1109/SENSORS56945.2023.10325204}}

\end{document}